\documentclass{article}

\usepackage{arxiv}

\usepackage[utf8]{inputenc} 
\usepackage[T1]{fontenc}    
\usepackage{url}            
\usepackage{booktabs}       
\usepackage{amsfonts}       
\usepackage{nicefrac}       
\usepackage{microtype}      
\usepackage{lipsum}		
\usepackage{graphicx}
\usepackage{doi}

\usepackage{amsmath}
\usepackage{multirow}
\usepackage{graphicx}
\usepackage{listings}

\usepackage{hyperref}

\usepackage{subcaption}

\title{A Deep Learning Method for Classification of Biophilic Artworks}

\author{\textbf{Purna Kar} \\
	Nottingham Trent University\\
	Nottinghamshire, United Kingdom \\
	\texttt{purna.kar@ntu.ac.uk} \\
	\And
 \textbf{Jordan J. Bird} \\
	Nottingham Trent University\\
	Nottinghamshire, United Kingdom \\
	\texttt{jordan.bird@ntu.ac.uk} \\
 	\And
 \textbf{Yangang Xing} \\
	Nottingham Trent University\\
	Nottinghamshire, United Kingdom \\
	\texttt{yangang.xing@ntu.ac.uk} \\
  	\And
 \textbf{Alexander Sumich} \\
	Nottingham Trent University\\
	Nottinghamshire, United Kingdom \\
	\texttt{alexander.sumich@ntu.ac.uk} \\
   	\And
 \textbf{Andrew Knight} \\
	Nottingham Trent University\\
	Nottinghamshire, United Kingdom \\
	\texttt{andrew.knight@ntu.ac.uk} \\
    	\And
 \textbf{Ahmad Lotfi} \\
	Nottingham Trent University\\
	Nottinghamshire, United Kingdom \\
	\texttt{ahmad.lotfi@ntu.ac.uk} \\
     	\And
 \textbf{Benedict Carpenter van Barthold} \\
	Vieunite Ltd.\\
	Birmingham, United Kingdom \\
	\texttt{benedict.carpentervanbarthold@vieunite.com} \\
}

\renewcommand{\shorttitle}{A Deep Learning Method for Classification of Biophilic Artworks}

\begin{document}
\maketitle

\begin{abstract}
Biophilia is an innate love for living things and nature itself that has been associated with a positive impact on mental health and well-being. This study explores the application of deep learning methods for the classification of Biophilic artwork, in order to learn and explain the different Biophilic characteristics present in a visual representation of a painting. Using the concept of Biophilia that postulates the deep connection of human beings with nature, we use an artificially intelligent algorithm to recognise the different patterns underlying the Biophilic features in an artwork. Our proposed method uses a lower-dimensional representation of an image and a decoder model to extract salient features of the image of each Biophilic trait, such as plants, water bodies, seasons, animals, etc., based on learnt factors such as shape, texture, and illumination. The proposed classification model is capable of extracting Biophilic artwork that not only helps artists, collectors, and researchers studying to interpret and exploit the effects of mental well-being on exposure to nature-inspired visual aesthetics but also enables a methodical exploration of the study of Biophilia and Biophilic artwork for aesthetic preferences. Using the proposed algorithms, we have also created a gallery of Biophilic collections comprising famous artworks from different European and American art galleries, which will soon be published on the Vieunite@ online community.
\end{abstract}

\keywords{Artificial neural networks \and Biophilia \and computer vision \and deep learning algorithms \and image classification \and intelligent buildings}

\section{Introduction}\label{sec:int}
The term ``\textit{Biophilia}" was first introduced by a psychoanalyst named Erich Fromm in 1973 in his book ``\textit{The Anatomy of Human Destructiveness}" \cite{5}. Biophilia is defined as a ``\textit{passionate love of life and of all that is alive... whether in a person, a plant, an idea, or a social group}", \textit{p. 406}. It was later popularised by socio-biologist E.O. Wilson in 1984 in his book ``\textit{Biophilia}" \cite{6}, where he defined it as ``\textit{the urge to affiliate with other forms of life}", \textit{p 416}. He proposed that human beings have an inherent tendency to connect to and thrive in natural environments and that exposure to nature or nature-inspired images positively impacts psychological well-being. 

To support this, exposure to natural environments was associated with elevated positive emotions and decreased negative emotions \cite{8}. Additionally, Biophilic designs have been shown to reduce stress, enhance creativity, improve productivity, and improve psychological well-being \cite{9,11,7}. Thus, with the rapid rise of urbanisation, it is pertinent to inspire design by nature in public places, especially healthcare facilities \cite{10}. The study \cite{10}, also claims exposure to Biophilic architecture has key benefits to people's mental health. Hence, architects and designers are encouraged to integrate Biophilic characteristics into the built environment,  and it has been argued that nature should be a fundamental component of any built structure rather than just an ``add-on” feature \cite{24}. A study of office workers shows this need to be close to nature \cite{23}. Compared to workers with office windows, those without windows were five times more likely to bring plants into their workspaces and three times more likely to bring pictures of nature. 

Art (e.g., painting, sculpture, architecture, visual design) inspired by nature or assimilating natural elements (whether or not an exact rendition of a natural scene, i.e. themes that resonate with nature or nature-inspired aesthetics) has been referred to as Biophilic art. Recent research has shown that art and cultural interpositions improve mental well-being. Some studies show frequent association with art results in overall well-being \cite{12}. Some authors propose that nature-inspired artwork reduces stress more effectively than direct exposure to the natural environment \cite{22}. Thus, there has been a call for the creation of ’immersive virtual experiences’ that leverage artistic vision and principles of psychological health, particularly for people living in urban environments who may face challenges in accessing natural settings \cite{26}. In such cases, artwork involving nature substitutes for full Biophilic environments and supports well-being in contexts with little scope of exposure to nature \cite{25}. Thus,  displaying Biophilic artworks in public places like stations, private spaces, hospitals, schools, and offices through digital display systems should become commonplace. The current study uses artificial intelligence to develop an art collection that can be used in a digital display system to make Biophilic art accessible. This collection can also be used for further studies by artists and researchers to understand the underlying intricacies that define Biophilic characteristics.

In the subsequent sections of the paper, relevant background studies and related work are discussed in Section \ref{sec:background}. The proposed method will be outlined in Sections \ref{sec:materials} and \ref{sec:theory}, Section \ref{sec:results} is dedicated to presenting and discussing our results. Finally, in Section \ref{sec:con}, we will discuss the key conclusions drawn from this research and the future scopes.

\section{Background and Related Work}\label{sec:background}
The categorisation of art in institutions such as museums and galleries is the preserve of highly trained specialists who build on the work of earlier curators and scholars. The process is time-consuming and relies on deep specialism and an immersive familiarity with art. This process might be substantially facilitated through the development of categorisation algorithms. The advent of deep neural networks \cite{21} particularly has revolutionised the computer vision domain by providing rapid, robust, and accurate solutions to categorisation problems \cite{17,18}. Deep neural networks are a sub-class of artificial neural networks that consist of several computational layers between the input and output layers. The intermediary layers are called hidden layers, and these are also referred to as ’deep’ layers. These networks are particularly compelling in solving complex problems like pattern recognition and feature extraction. Such networks comprise several layers like the input, convolutional, activation, pooling, fully connected, and output layers, with each layer designed to perform a particular task of extracting features, reducing spatial dimensions, and learning high-level features and relationships. These networks need to be trained on properly labeled datasets through backpropagation. During the training cycle, the network adjusts its learnable parameters based on the loss function, i.e. a function based on the difference between predicted and actual values, to improve its predictive accuracy. All deep learning models require rigorous training on properly annotated datasets, hence the performance of the model depends on the quality of data the model trains on. Needless to say, using large and accurately labeled data is crucial to the success of any classification model.

Currently, the categorisation of paintings is mostly done by theme, style, artist, or movement. The categories are often interpretative and can overlap or intertwine with one another. Thus, building a model that can accurately predict the class is difficult, and models fail to learn the underlying patterns and generalise for each trait accurately. In \cite{16}, a deep learning model was trained on several works by famous artists, categorised with respect to brushstrokes. Other models have been trained to classify based on genres, styles, and schools of art. For example, an algorithm described by \cite{13}, extracts global and local features from images to classify paintings into different styles. The global feature comprises colour statistics, and the local features consist of features extracted from the image segmentation using colour features. Finally, classification is performed using a self-organising map (SOM) as proposed by Kohonen \cite{30}. Other authors fine-tuned convolutional neural networks (CNN) to perform classification based on style, artist, genre, period, and association with national artistic context \cite{14}, which achieved an overall accuracy of 75\%-80\%. In another study \cite{15}, the authors conducted a comparative study of various classification methodologies that included a two-stage process. In the first phase, the images are split into five patches and a deep neural network is used for the classification of the patches independently; the results are in the form of probability vectors. The second stage uses a shallow neural network to predict the genre by considering the results of the previous stage. This method achieved an accuracy of 96\%. During classification, deep learning techniques have been used to create well-labelled art collections \cite{20} where the authors have created a dataset that comprises digital paintings from 91 different artists.

Much research has been done to perform the classification of artworks based on parameters such as art style, genre, artist, period, etc. However, to the authors’ knowledge, no published research categorises artworks based on Biophilic attributes. The current study aims to build a model that performs the classification of artworks based on Biophilic features (see Section \ref{subsec:bio} for feature details). Specifically, a deep learning model is trained that is capable of predicting Biophilic characteristics in art in a way that can facilitate the curation of a Biophilic gallery of famous pieces. Such a model would help public galleries and digital communities organise a large collection of artworks and segregate paintings that are Biophilic. In addition, developing such a classification system would benefit research in computer vision, and other domains (e.g., art, psychology) that study the impact of nature-inspired environments on health and well-being.

\section{Materials and Methods}\label{sec:materials}
Our study has two primary goals: i) to develop a pipeline capable of generating Biophilic labels from digital copies of paintings and to define the dominant label amongst the predicted labels; ii) to use the labels to create an organised gallery of Biophilic artwork that can be readily used by galleries, other public places and in research.

\subsection{Data and Resources}\label{subsec:data}
To the authors’ knowledge, no publicly available dataset consisting of artworks and the corresponding Biophilic labels exists, and no previous study uses AI to classify art based on Biophilic traits. For the current study, a fully annotated dataset was created from scratch by scraping online art galleries for digital copies of images. Almost all well-known galleries have high-resolution digital copies of all famous works. Five thousand digital copies of paintings were downloaded from galleries, including ’The Chicago Art Institute’, ’The Philadelphia Museum of Art’, ’The Getty Center’, ’The Metropolitan Museum of Art’, ’The National Gallery of Art’, ’Rijksmuseum’, ’Nasjonalmuseet’, and ’The Smithsonian American Art Museum’. Only CC0 or public domain images were included in the set. An additional 60 AI-generated images were included from the Vieutopia platform. Several augmentation/transformation techniques were used to increase the dataset size to 15,097, such as horizontal flip, shear of +/- 11 degrees and +/- 6 degrees vertical, rotation between -8 degrees and +8 degrees, brightness between -25\% and +25\%, blur up to 1.5px, and added noise up to 2\% of pixels. These techniques help increase the diversity of the dataset by including variations that reflect real-world scenarios, making our deep learning model robust, which improves model learning for underlying patterns and generalisability. Data augmentation can also address several other problems, like class imbalance, a common issue in multi-class datasets with unequal representation of classes. Figure \ref{fig:distrib} shows our dataset's overall distribution of classes. It shows that some classes are hugely under-represented and data augmentation techniques can help in generating multiple synthetic samples of the under-represented classes. Data Augmentation also helps reduce overfitting problems by acting as a regularisation. The dataset comprises a folder containing the digital artwork and a CSV file that records whether or not a class is present for each image. The columns represent 15 classes, and the rows represent each image. For building the deep learning models, we have used Pytorch libraries \cite{19},  and for explainable AI, we used Local Interpretable Model-agnostic Explanations (LIME) \cite{4}.

\subsection{Biophilic Labels}\label{subsec:bio}
\begin{figure}
\centering
\includegraphics[width=1.0\linewidth]{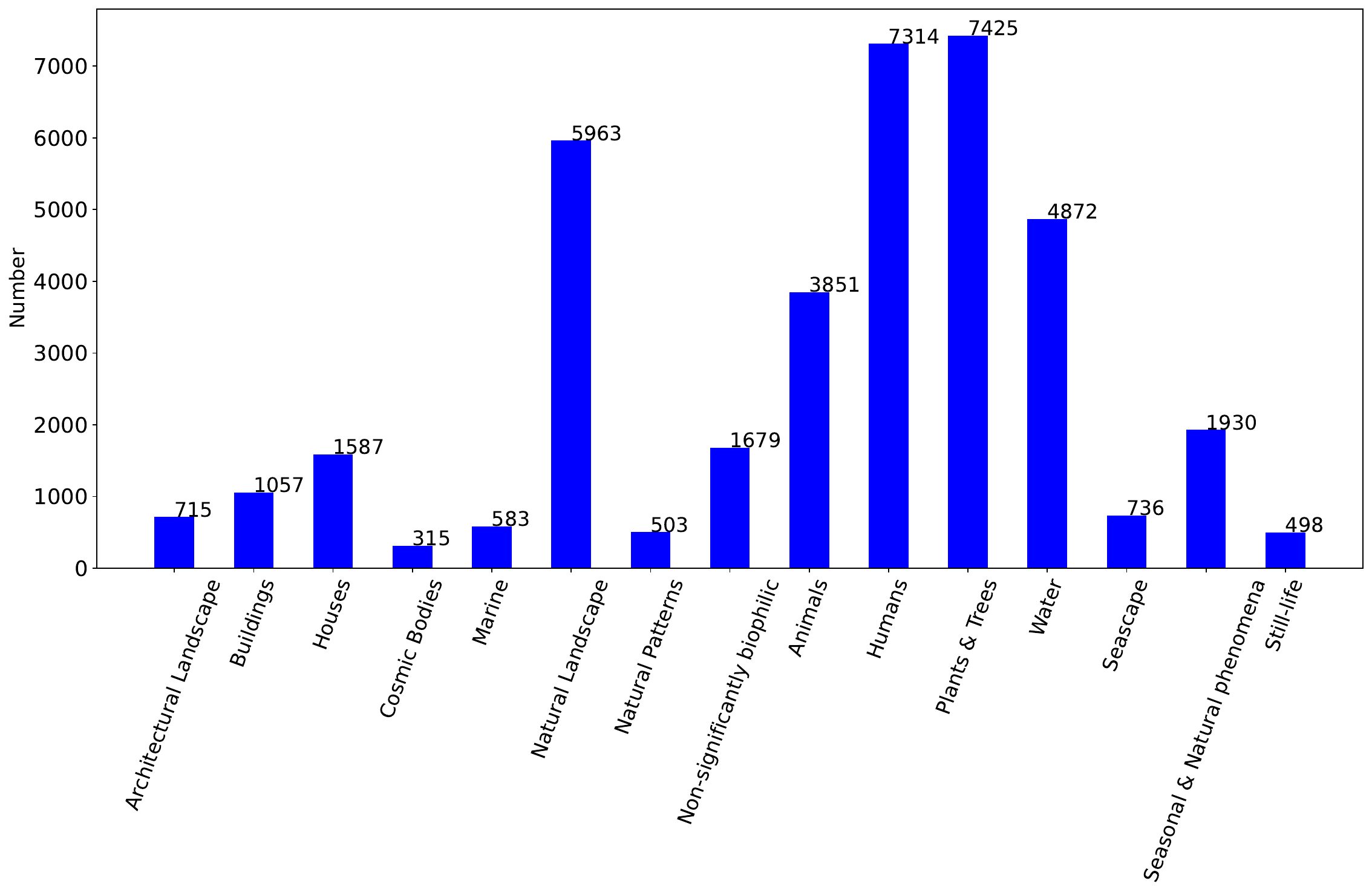}
\caption{\label{fig:distrib}Distribution of Biophilic labels in the Dataset.}
\end{figure}

There is no known technique for classifying artworks based on Biophilic features; therefore, no standard labels for Biophilic design in art are defined. The most established Biophilic characteristics for architectural designs have been stated in \textit{Nature Inside: A Biophilic design guide} \cite{31}, \textit{p5}; these features have been developed in three general categories:
\begin{itemize}
\item \textit{Nature in the space} with subcategories: Visual Connection w/ Nature, Non-Visual Connection with Nature, Non-Rythmic Sensory Stimuli, Thermal \& Airflow Variability, Presence of Water, Dynamic \& Diffuse Light, and Connection with Natural Systems.
\item \textit{Natural Analogues} with subcategories: Biomorphic Forms \& Patterns, Material Connection with Nature, and Complexity \& Order.
\item \textit{The Nature of the Space} with subcategories: Prospect, Refuge, Mystery, Risk/Peril, and Awe
\end{itemize}
There are fundamental differences between architectural design and visual art, not least the spatial and temporal manner in which the built environment is experienced relative to a static painting and the broader range of sensory stimuli that pertain to the experience of architecture. Moreover, several features of \textit{The Nature of Space} category have an emotional quotient, which is ambiguous and open to interpretation. For these reasons, we develop categories based on nameable objects commonly depicted in paintings. The Fifteen different Biophilic classes that were used are as follows:
\begin{itemize}
\item \textit{Architectural Landscape}: Physical aspects of the built environment, e.g., buildings and structures, historical ruins, cityscapes, and gardenscapes.
\item \textit{Buildings}: Depicting multistory buildings.
\item \textit{Houses}: Depicting houses, huts, and other vernacular home structures
\item \textit{Cosmic bodies}: Depicting celestial objects like the Sun, the Moon, Planets, Galaxies, Comets, etc.
\item \textit{Marine}: Paintings of the sea with ships or warships as the main subject.
\item \textit{Natural landscape}: Depicting natural scenery with hills, mountains, rivers, forests, deserts, etc.
\item \textit{Natural patterns}: Biomorphic shapes or shapes and patterns occurring naturally, e.g., fractal rings, patterns in waves or leaves.
\item \textit{Animals}: Presence of any animal.
\item \textit{Humans}: Presence of humans or depiction of humans either portrait or integration with nature.
\item \textit{Plants \& Trees}: Plants, trees, bushes, herbs, etc.
\item \textit{Water}: Presence of water in any form, e.g., sea, river, lake, pond, waterfall, etc.
\item \textit{Seascape}: Depicting sea beaches, cliffs, and sea view
\item \textit{Seasonal \& Natural phenomena}: Seasonal and weather reflections or any natural phenomena like storms, volcanic eruptions, rainbows, etc.
\item \textit{Still life}: Still life paintings of fruit, shells, books, and other small `table-top' objects.
\item \textit{Non-significantly Biophilic}: Neutral or non-Biophilic paintings.
\end{itemize}
The class: \textit{Seasonal \& Natural phenomena} have been further divided into 5 classes:
\begin{itemize}
\item \textit{Natural Phenomena}: Volcanic eruptions and rainbows.
\item \textit{Autumn}: Autumn or Fall scenes.
\item \textit{Winter}: Winter and snowy scenes.
\item \textit{Stormy weather}: depicting rainy, storms, hurricanes.
\item \textit{Northern lights}: depicting Aurora Borealis.
\end{itemize}
Roboflow environment \cite{32} was used for manually annotating each image with Biophilic labels (each image could have more than one label). Figure \ref{fig:distrib} shows the distribution of Biophilic labels in the images. As can be seen, the dataset is unbalanced, and some labels are underrepresented.

\subsection{Classification Task}
Our classification model is designed to operate by taking a digital copy of an artwork as input and providing the Biophilic labels (discussed in Section \ref{subsec:bio}) as well as discerning the most significant Biophilic label within the artwork. We have used state-of-the-art computer vision algorithms for extracting meaningful Biophilic features from the artwork, which assist us in categorising images based on their Biophilic characteristics by accentuating the dominant Biophilic features that contribute to a deeper understanding of the natural world captured within the artwork and helps in curating a gallery of high-quality nature-inspired artworks. The details of the model have been discussed in Section \ref{sec:theory}.

\subsection{Training and Validation}
The dataset collected for this study is divided into the ratio of 7:2:1 for training, validation, and testing respectively. Therefore, 10869 images were used for training, 2718 for validation, and 1510 for testing. As described in Section \ref{subsec:data}, data augmentation techniques were used to increase the size of the dataset and balance underrepresented categories. Figure \ref{fig:distrib} shows the overall distribution of the classes in the dataset. We have performed hyperparameter tuning to select the best hyperparameter for our deep learning model; for this purpose, we have used Optuna \cite{28}. For each hyperparameter, the following spaces have been defined:
\begin{itemize}
\item`Optimizer': `Adam' and `SGD'
\item`Learning Rate': 0.001, 0.002, 0.003, 0.004, 0.005, 0.006, 0.007, 0.008, 0.009 and 0.01
\end{itemize}
We conducted 10 trials, each with 5 epochs for each hyperparameter value, and the best results were achieved by a learning rate of 0.001 and Adam optimizer \cite{29}. At the beginning of the training process, image preprocessing was performed to resize to (224x224), normalise, convert to a tensor, and load to a training loader (along with the corresponding labels) using batch sizes of 12. The labels for each image are a binary array of size 15, which represents whether a category is present or not, with 0 representing absence and 1 representing presence. The deep learning algorithm is trained on the training dataset in batches of 12. Once the model is trained on all batches of the images from the training dataset, the model is used to predict the labels of images using the validation dataset. We compare the results of the predicted labels with the actual labels on the validation dataset and note the classwise F1 score. The training and validation cycle is continued for 50 Epochs, and the model with the best F1 score on the Validation dataset is saved.

\section{Overview of Proposed Approach}\label{sec:theory}

Image classification refers to algorithms designed to assign one or more labels to an image. There are three types of classification: binary, multi-class, and multi-label. For binary and multi-class classifications, the algorithm outputs a single category. In multi-label classification, there can be multiple categories. Most images have complex and diverse scenarios where multiple objects or concepts of interest may be present, and thus, a multi-label classification is preferred. The current study sought to find all Biophilic labels present in any given artwork. Hence, a multi-label classification technique has been developed that is capable of predicting all the Biophilic classes that are present in each piece. On the other hand, object detection is widely used in solving computer vision problems where instances of semantic objects of pre-defined categories are identified and localised in an image. Some of the Biophilic labels incorporated are concepts, such as ’season’ and ’natural phenomena’, that cannot be identified by an object detection algorithm. Furthermore, the localisation of objects was not anticipated. Thus,  a multi-label classification algorithm is used to find all the Biophilic traits in the image.

We have used an encoder-decoder model to perform our multi-label image classification task. An encoder-decoder model comprises two parts: (i) an \textit{encoder}, where an input image is represented in a lower dimension, and (ii) a \textit{decoder}, which uses the image representation to predict the labels present. The task of the encoder is to take an input image and extract features by passing through several convolutional layers. The convolutional layers are built using multiple filters using different kernel sizes capable of extracting features from an image at different levels of abstraction and producing a final vector representation of the image of a much lower dimension. The current study used OpenAI’s Contrastive Language-Image Pretraining (CLIP) image encoder, Vision Transformer (ViT)-B/32 \cite{3}, to generate the embeddings of each image. CLIP or Contrastive Language-Image Pretraining \cite{3} CLIP was developed by OpenAI to apprehend texts and image pairs. 

The contrasting learning objective of CLIP is based on embedding similarities. The similarity is measured via cosine similarity:
\begin{equation}
    \text{sim}(v_i, w_j) = \frac{v_i \cdot w_j}{\|v_i\| \|w_j\|}
\end{equation}

\noindent where $v_i$ and $w_j$ are the two input vectors. $\text{sim}(v_i, w_j)$ denotes dot product, and where $\|v_i\|$ and $\|w_j\|$ are the Euclidean lengths. Cosine similarity is used in CLIP's contrastive loss function, which is given as:
\begin{equation}
    L = -\log \frac{\exp(\text{sim}(v_i, w_i) / \tau)}{\sum_{j=1}^{N} \exp(\text{sim}(v_i, w_j) / \tau)}
\end{equation}

\noindent where $\tau$ is a temperature scaling parameter to control the distribution of similarity scores. The optimisation of contrastive loss thus enables the learning of image embeddings that are both visually and semantically descriptive. In the original implementation, $\tau = 0.07$ \cite{3}.

\begin{figure}[]
    \centering
    \includegraphics{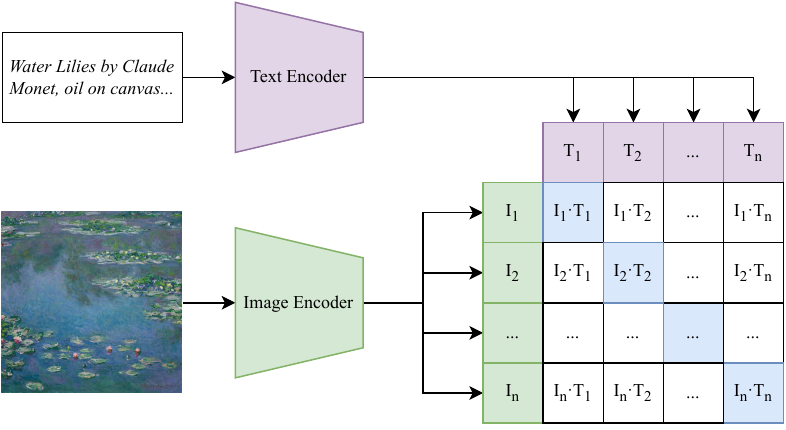}
    \caption{A diagram of contrastive pre-training, adapted from \cite{3}.}
    \label{fig:clip-diagram}
\end{figure}

\begin{figure}[]
    \centering
    \includegraphics{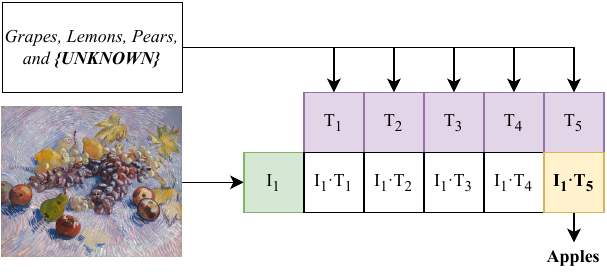}
    \caption{Zero-shot prediction example using CLIP embeddings, adapted from \cite{3}.}
    \label{fig:clip-diagram-zeroshot}
\end{figure}

An example of contrastive pre-training can be seen in Figure \ref{fig:clip-diagram}. Zero-shot predictions for images can be made using CLIP embeddings as shown by Figure \ref{fig:clip-diagram-zeroshot}.

The image encoder of CLIP is built on the Vision Transformer (ViT) \cite{27} architecture. A ViT splits an image into patches of a predefined shape and arranges them into a linear vector. Next, multiple transformer blocks extract patterns from this vector to generate image embeddings. The Vit-B/32 transforms an image to a unidimensional array of length 512. The CLIP image encoder is pre-trained on 400 million images of over a thousand classes. Hence, it is considered to be the state-of-the-art feature extraction algorithm. For the decoder, three linear layers were used; the first two linear layers are followed by batch normalisation and ReLU activation functions, whereas the last layer is followed by a Sigmoid activation function only. The last layer outputs an array of length 15, each item in the array is a probability of the occurrence of a Biophilic label. Batch Normalisation involves normalising the output of each layer in mini-batches. This regularisation technique ensures a consistent distribution of activations, thus improving the stability and speed during the training process. ReLU or Rectified Linear Unit is an activation function that injects non-linearity in deep learning networks, which helps the networks to learn complex patterns. The Sigmoid function is another activation function given by the formula: 
\begin{equation}
sigma(x) = \frac{1}{(1+e^{-x})}
\end{equation}
$sigma(x)$ is the output of the sigmoid function for an input $x$. 
This function maps any real number to the range of $[0,1]$. Our output is a 15-size array (representing all the 15 Biophilic labels) that depicts the probability that each class is present in the image; therefore, using the Sigmoid function as the last layer in the output ensures that all the outputs are within the range [0,1]. Another feature of the Sigmoid function is that this activation function treats each output independently; that is, the probability of a particular class is independent of whether any other class is present. Using the Sigmoid function converts our classification problem to n-binary classification problems.  This makes Sigmoid the best for multi-label classification models as we are interested in getting all labels present in an image, and the presence of one class does not affect the presence of any other classes. The summary of the decoder model can be seen in Figure \ref{fig:Decoder}

\begin{figure}
    \centering
    
    \begin{lstlisting}[language=Python]
    
    Decoder(
        (fc1): Linear(int_features=512, out_features=256, bias=True)
        (batch_norm1): BatchNorm1d(256, eps=1e-05, momentum=0.1, affine=True,
        track_running_stats=True)
        (relu1): ReLu()
        (dropout1): Dropout(p=0.2, inplace=False)
        (fc2): Linear(int_features=256, out_features=128, bias=True)
        (batch_norm2): BatchNorm1d(128, eps=1e-05, momentum=0.1, affine=True,
        track_running_stats=True)
        (relu2): ReLu()
        (dropout2): Dropout(p=0.2, inplace=False)
        (fc3): Linear(int_features=128, out_features=15, bias=True)
    )
    \end{lstlisting}
    \caption{Summary of the Decoder model.}
    \label{fig:Decoder}
\end{figure}

 During the training and validation phase, we used Adaptive Moment Estimation (Adam) \cite{29}, which is an optimisation technique most commonly used for training deep learning models. This algorithm is designed to update the learnable parameters of the network by minimising the loss function during the training process. This technique uses two optimisation algorithms: RMSprop and Stochastic Gradient Descent with Momentum to conserve the learning rate for each learnable parameter. The loss function used in the training phase is Binary Cross Entropy (BCE),  which is computed for each batch, where the loss gradients of all tensors are computed, following which the weights of all the learnable parameters are updated. The last layer of the Decoder model is the Sigmoid Activation function, which converts the classification problem to an n-binary classification problem, which defines the probability of presence for each class in any given image. Hence, we use the BCE Loss, to minimise loss to update the weights and biases of the network. The BCE loss is given by the formula: 
\begin{equation}
L_{BCE} = \frac{-1}{n}\sum_{i}^{n} {Y_i.logP_i+(1-Y_i).log(1-P_i)}
\end{equation}
Where Y is the actual value and P is the predicted value. 
For evaluating our model we use the following metrics: Precision, Recall \& F1 score. Precision is the positive predictive value and is given by the formula,
\begin{equation}
Precision = \frac{TP}{TP+FP}
\end{equation}
Recall also known as sensitivity or true positive rate, is given by the formula, 
\begin{equation}
Recall = \frac{TP}{TP+FN}
\end{equation}
F1 score uses both Precision and Recall and is defined as  
\begin{equation}
F1 score = \frac{2*(Precision * Recall)}{Precision + Recall}
\end{equation}
TP, FP, TN, and FN are the True Positives, False Positives, True Negatives, and False Negatives. High precision signifies that the model is capable of returning more relevant results than irrelevant ones, and it is a measure of quality. Whereas high recall indicates that the model is capable of returning all relevant results, it is a measure of quantity. The F1 score tells us how well-performing a model is, it shows the model is capable of recognising positive cases and minimises the false positive and false negative cases.

\section{Results and Discussion}\label{sec:results}

\subsection{Performance of Model}

\begin{figure}[]
    \centering
    \begin{subfigure}[b]{0.45\textwidth}
        \centering
        \includegraphics[width=\textwidth]{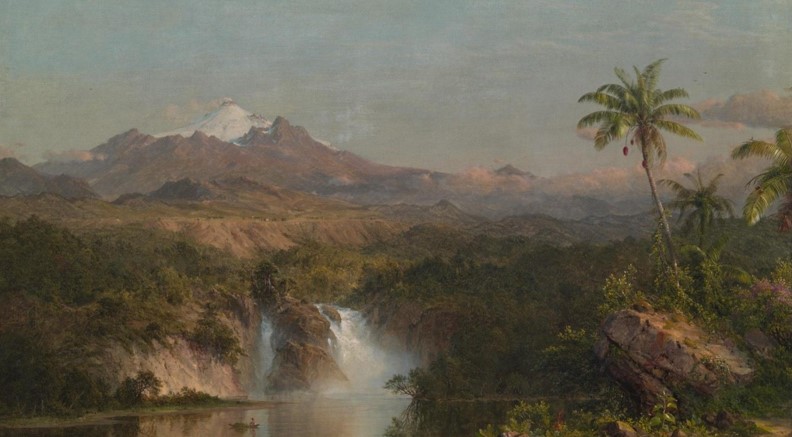}
        \caption{Natural Landscape, Plants \& Trees, Water}
        \label{fig:first-left}
    \end{subfigure}
    \hfill 
    \begin{subfigure}[b]{0.45\textwidth}
        \centering
        \includegraphics[width=\textwidth]{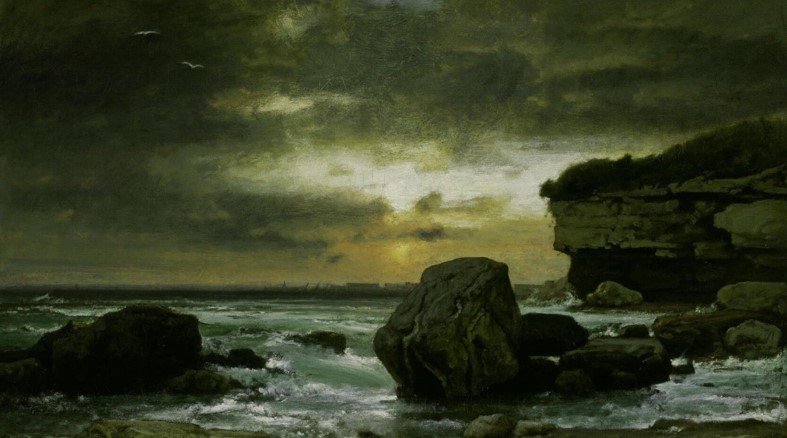} 
        \caption{Water, Seascape, Stormy Weather}
        \label{fig:first-right}
    \end{subfigure}
    \begin{subfigure}[b]{0.45\textwidth}
        \centering
        \includegraphics[width=\textwidth]{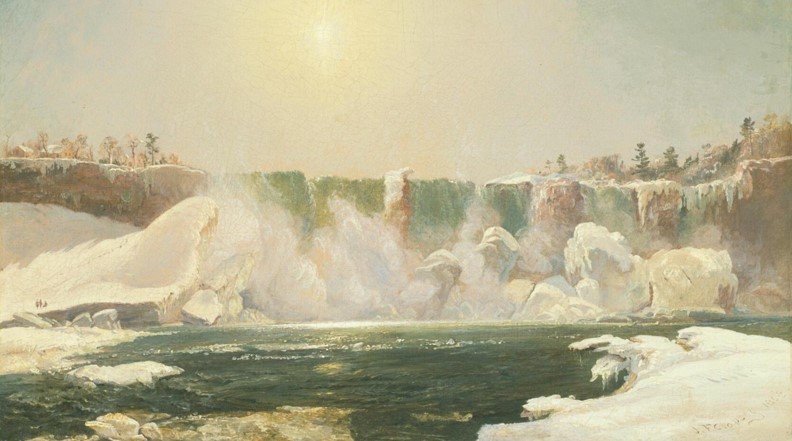} 
        \caption{Natural Landscape, Water, Winter}
        \label{fig:second-left}
    \end{subfigure}
    \hfill 
    \begin{subfigure}[b]{0.45\textwidth}
        \centering
        \includegraphics[width=\textwidth]{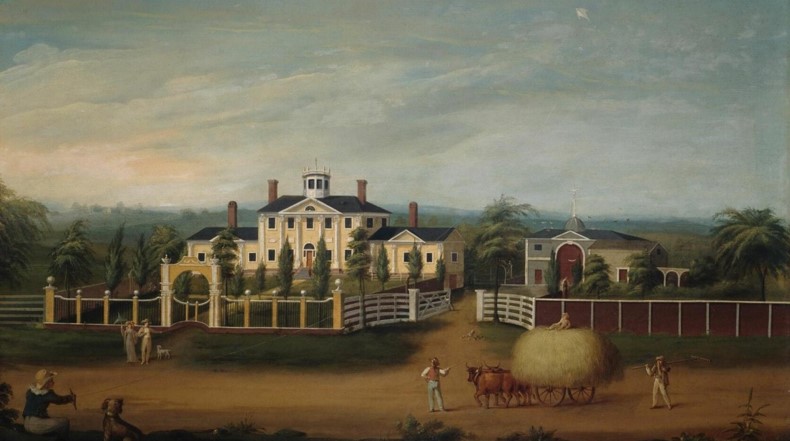} 
        \caption{Architectural Landscape, Buildings, Animals, Humans, Plants \& Trees}
        \label{fig:second-right}
    \end{subfigure}
    \begin{subfigure}[b]{0.45\textwidth}
        \centering
        \includegraphics[width=\textwidth]{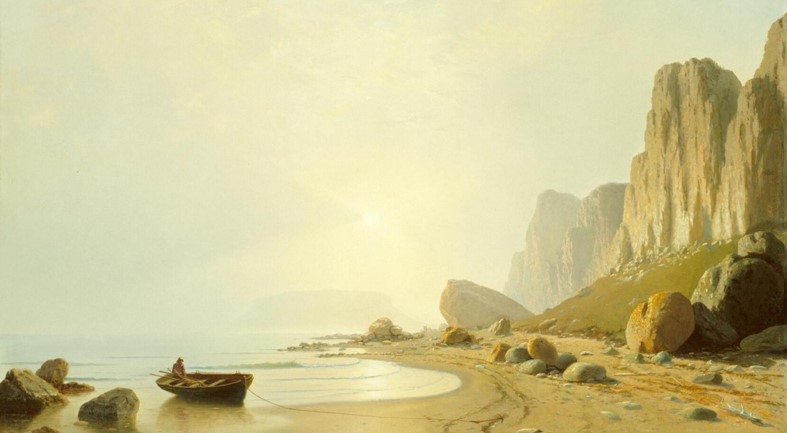} 
        \caption{Humans, Water, Seascape}
        \label{fig:Third-left}
    \end{subfigure}
    \hfill 
    \begin{subfigure}[b]{0.45\textwidth}
        \centering
        \includegraphics[width=\textwidth]{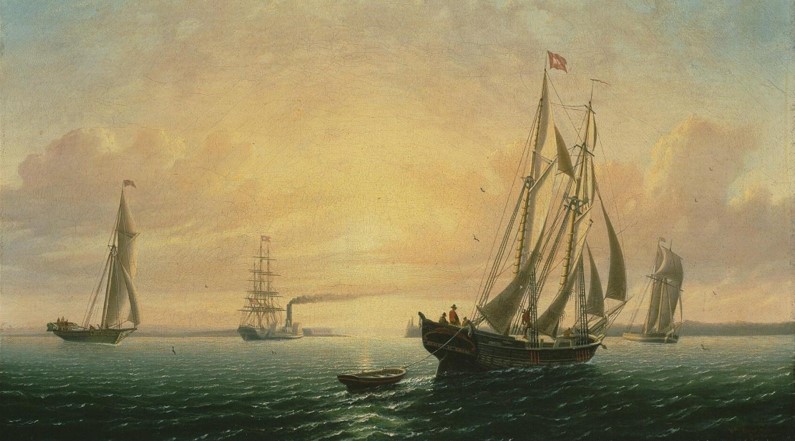} 
        \caption{Marine, Humans, Water}
        \label{fig:Third-right}
    \end{subfigure}
    \begin{subfigure}[b]{0.45\textwidth}
        \centering
        \includegraphics[width=\textwidth]{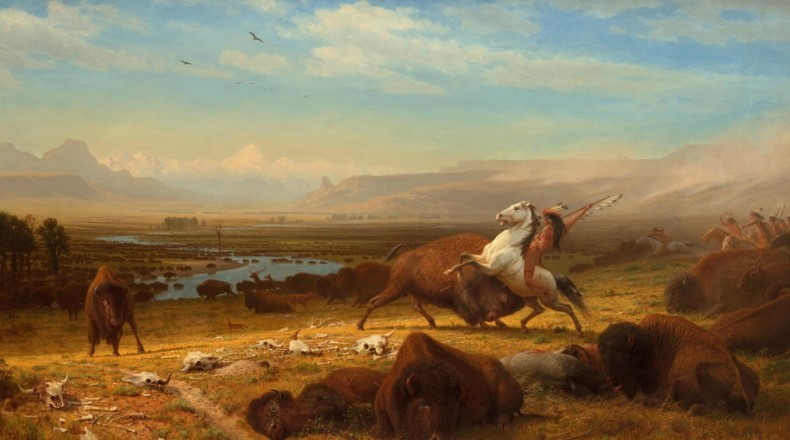} 
        \caption{Natural Landscape, Animals, Humans}
        \label{fig:fourth-left}
    \end{subfigure}
    \hfill 
    \begin{subfigure}[b]{0.45\textwidth}
        \centering
        \includegraphics[width=\textwidth]{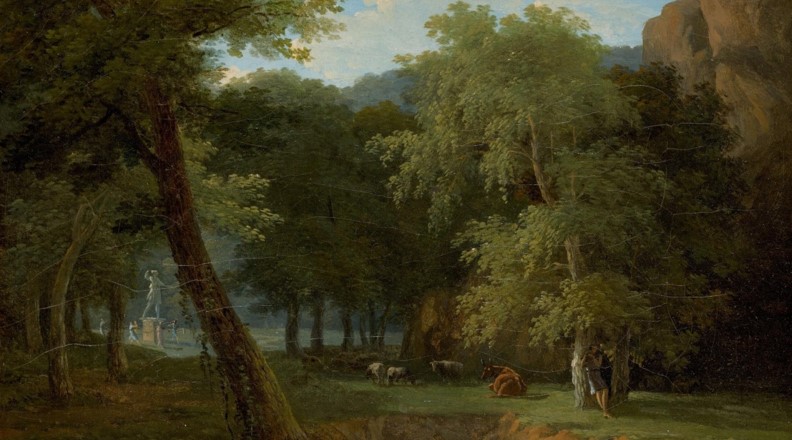} 
        \caption{Natural Landscape, Animals, Humans, Plants \& Trees}
        \label{fig:fourth-right}
    \end{subfigure}
    \caption{Random examples of images from the dataset along with classification predictions.}
    \label{fig:sample}
\end{figure}

We load the best model from the Training-Validation step and use it to predict the categories for the images in the test dataset. For each image, the output is an array of size 15, which represents the probability of occurrence of each category. We use a threshold of 0.5 to determine whether a category is relevant or not, i.e. if the probability of the category is greater than 0.5 then we assume the category is present, or else it is absent. In this way, we convert the output to a binary array of size 15. Since our dataset is imbalanced, accuracy is not the most appropriate metric, instead we use three other metrics, namely Precision, Recall, and F1-score to evaluate our model. Tables \ref{tab:Report}, and \ref{tab:Report2} show the key classification metrics of our model on the test or unseen data. The weighted F1 score on the test dataset is 90.61\%. Figures \ref{fig:sample}, show some of our best results; as can be observed, the model has a strong ability to identify all the Biophilic labels present in the artworks. The algorithm also predicts the most dominant label along with the Biophilic labels. This is done by finding the Biophilic label that has the highest probability of occurrence among the 15 labels.

\begin{table}
\centering
\caption{\label{tab:Report}Classification Report of the performance of Model on Test Data for each Biophilic Category.}
\begin{tabular}{llll}
\hline
Biophilic Labels & precision & recall & f1-score \\\hline
Architectural Landscape & 0.83 &  0.89 & 0.86 \\
Buildings & 0.91 &  0.82 & 0.86 \\
Houses & 0.85 &  0.81 & 0.83 \\
Cosmic Bodies & 0.89 &  0.62 & 0.73 \\
Marine & 0.87 &  0.95 & 0.91 \\
Natural Landscape & 0.88 &  0.95 & 0.91 \\
Natural Patterns & 0.94 &  0.72 & 0.81 \\
Non-significantly Biophilic & 0.97 &  1.00 & 0.99 \\
Animals & 0.91 &  0.78 & 0.84 \\
Humans & 0.90 &  0.93 & 0.92 \\
Plants \& Trees & 0.96 &  0.94 & 0.95 \\
Water & 0.86 &  0.93 & 0.89 \\
Seascape & 0.97 &  0.83 & 0.89 \\
Seasonal \& Natural phenomena & 0.91 &  0.94 & 0.93 \\
\hline

Weighted F1-score & & & 0.9062 \\
\hline
\end{tabular}

\end{table}

\begin{table}
\centering
\caption{\label{tab:Report2}Overall performance of the Model}
\begin{tabular}{llll}
\hline
Metric (average) & precision & recall & f1-score \\\hline
Micro & 0.91 &  0.91 & 0.91 \\
Macro & 0.91 &  0.87 & 0.89 \\
Weighted & 0.91 &  0.91 & 0.91 \\
Samples & 0.92 &  0.93 & 0.91 \\
\hline
\end{tabular}
\end{table}

\subsection{Comparison With Other Models}
Performance was compared with two other models, a ResNet50 model \cite{1}, pre-trained on ImageNet dataset from the Pytorch library, and a Data-efficient Image Transformer (DEIT) transformer (deit-tiny-patch16-224) \cite{2} pre-trained on ImageNet-1k. A ResNet-50 \cite{1} is one of the most popular deep neural networks used in image classification tasks. It has a depth of 50 layers and uses a residual block to build the architecture. A residual block consists of several convolutional layers where the input of a block is combined with the output using skip connections. This skip connection makes it easier for the network to train by learning the difference or residual between the output and the input. The skip connections of the ResNet also help diminish the vanishing gradient problem by allowing the gradients to propagate to the initial layers with greater magnitude by skipping a few in-between layers. Vanishing gradients, if not addressed, can lead to overfitting and potentially suboptimal results. A DEIT \cite{2} is inspired by transformer-based architectures designed to perform Natural Language Processing (NLP). A DEIT has been specifically designed to achieve high performance using smaller datasets. The DEIT model uses transfer learning through a technique called \textit{distillation through attention} and tokenises the images by creating a sequence of patches that do not overlap for processing by the transformer.

Table \ref{tab:Accuracy} tabulates the performance of the three methods. We used the weighted F1 score as an evaluation metric for our comparison. As can be seen, using an encoder-decoder model outperforms the other two models. DEIT, a transformer-based model, performs poorly in capturing detailed spatial information. Additionally, the tokenisation process increases complexity and overhead. The image encoder used in our model uses the Open AI CLIP image embeddings \cite{3} that have been trained on 400 million images belonging to more than a thousand classes, whereas ResNet50 is trained on the ImageNet dataset which only has a million images. Thus, CLIP is often more efficient in extracting features from images than the ResNet50 model \cite{tu2024closer,fan2024improving}.

\begin{table}[]
\centering
\caption{\label{tab:Accuracy}Accuracy of different models.}
\begin{tabular}{l|r}
\hline
Method & Weighted F1 score (\%) \\\hline
Our Method & 90.62 \\
ResNet50 & 86.15 \\
DEIT & 82.19 \\
\hline
\end{tabular}
\end{table}

\subsection{Explainable AI}
\begin{figure}[]
    \centering
    \begin{subfigure}[b]{0.49\textwidth}
        \centering
        \includegraphics[width=\textwidth]{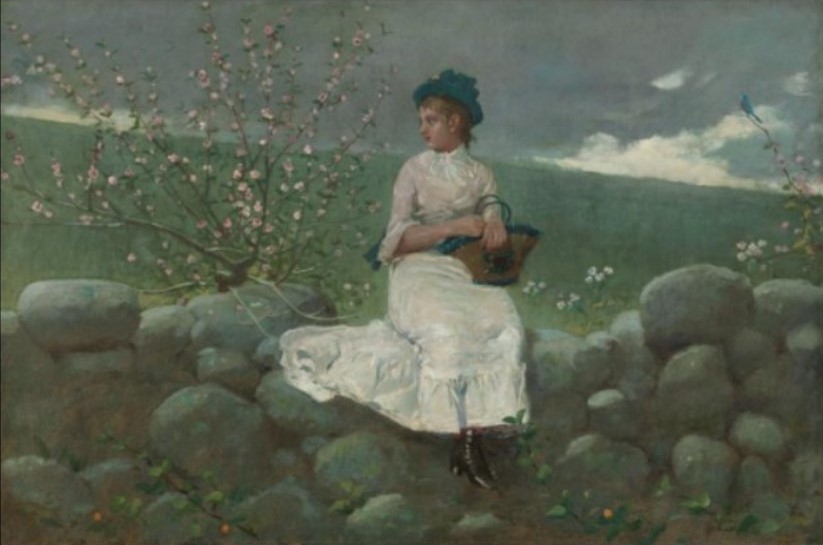}
        \caption{Predicted Labels: Humans, Plants \& Trees}
        \label{fig:top-left}
    \end{subfigure}
    \hfill 
    \begin{subfigure}[b]{0.49\textwidth}
        \centering
        \includegraphics[width=\textwidth]{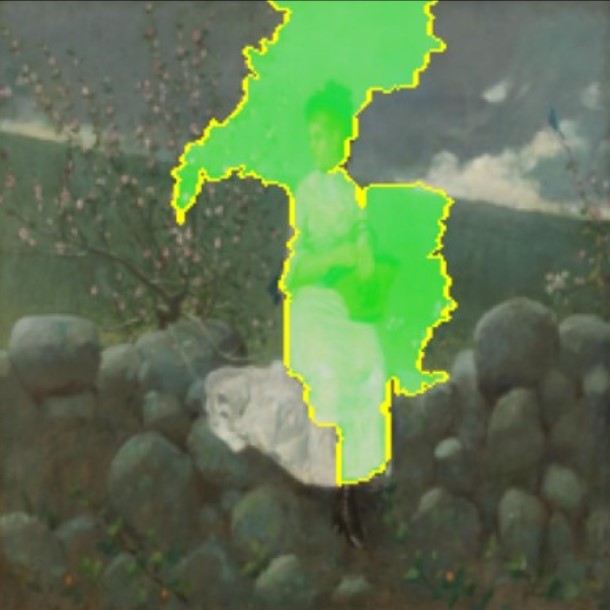} 
        \caption{LIME: Humans}
        \label{fig:top-right}
    \end{subfigure}
    \begin{subfigure}[b]{0.49\textwidth}
        \centering
        \includegraphics[width=\textwidth]{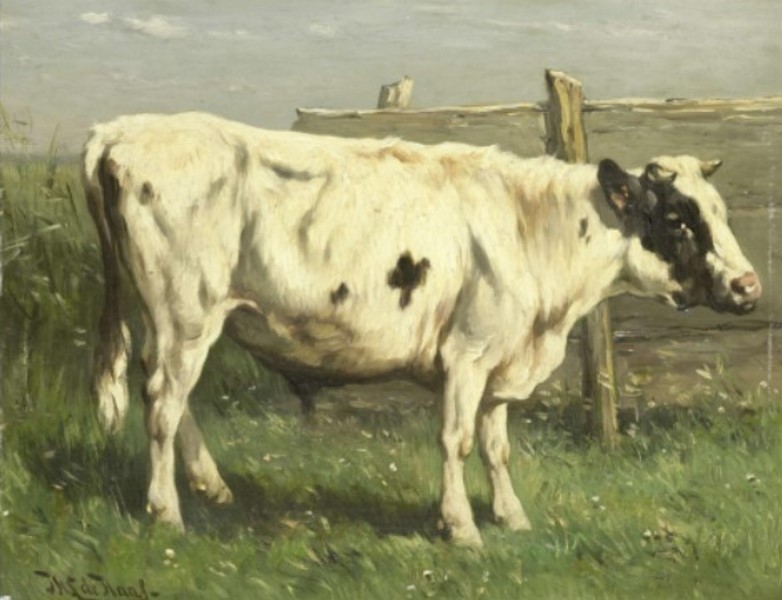} 
        \caption{Predicted Label: Animals}
        \label{fig:bottom-left}
    \end{subfigure}
    \hfill 
    \begin{subfigure}[b]{0.49\textwidth}
        \centering
        \includegraphics[width=\textwidth]{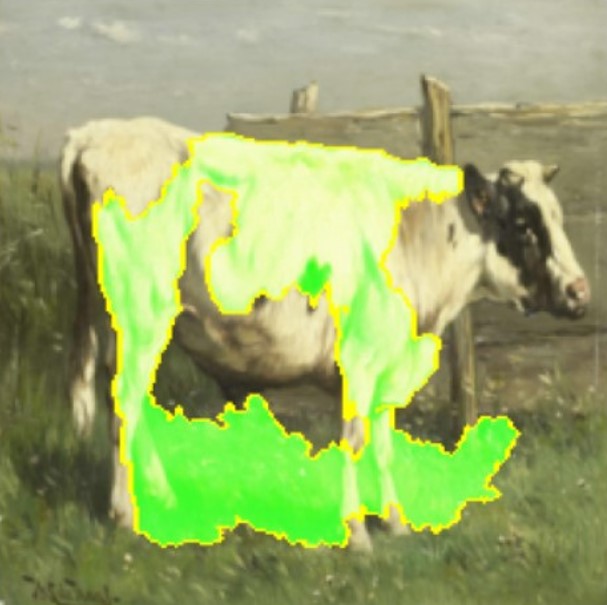} 
        \caption{LIME: Animals}
        \label{fig:bottom-right}
    \end{subfigure}
    \caption{Examples of explainable model predictions from our approach using the LIME algorithm.}
    \label{fig:explain}
\end{figure}

Deep neural networks have a very complex architecture and are often referred to as a ``Black Box". This makes it very difficult for users to understand the logic and mechanisms behind the output decisions. From the \textit{black box}, predictions are made following a series of calculations that are so complex that they are typically beyond human ability to visualise or understand. Explainable AI capabilities assist users in understanding and interpreting the decision-making logic of the deep learning model. Most deep learning models are made up of a large number of convolutional layers, which can be difficult to interpret. Explainable AI algorithms approximate such a model with simpler models for ease of interpretation. To explain the current model, Local Interpretable Model-agnostic Explanations (LIME) \cite{4} were used to prioritise the interpretation of individual results of the model by creating simpler surrogate models that give a coarse-grained interpretation of the deep learning network. Initially, surrogate instances are created by spawning a series of perturbed instances around the point of interest. The perturbed instances are created by randomly changing the features by slight measures. Then, a much simpler and interpretable surrogate model (e.g. decision tree, linear, or polynomial regressor model) is trained using these instances, along with the deep learning prediction for the perturbed instances. This way, the surrogate model can interpret the deep learning model logic locally and help users gain valuable insights into the mechanisms of the trained deep learning model. Figures \ref{fig:explain} show visual examples of Explainable AI, highlighting regions that contributed towards the model's decision to predict the most dominant label. The regions highlighted in green represent the areas that influenced the model in predicting the most likely output.

\subsection{Curation of Biophilic Gallery}

The current study included a collaboration with Vieunite (Allsee)  to curate a gallery of Biophilic artwork. Vieunite, which pairs the Vieunite art library with its innovative Textura digital canvas, is a platform for art lovers and artists.  Using the current Biophilic classifier, digital artworks were labeled by generating Biophilic tags. The trained classifier was used to generate Biophilic tags and the most dominant Biophilic class for each artwork in the Vieunite platform. A threshold of 0.65 (65\%) was used so that only the confident predictions are considered in the tag generation process. The artwork is also provided with an additional label called ’Biophilic flag’ which states whether the artwork is Biophilic in nature or not significantly Biophilic. If the predicted Biophilic labels have any of the following labels in them: ’Autumn’, ’Natural Phenomena’, ’Northern Lights’, ’Stormy Weather’, ’Winter’, ’Cosmic Bodies’, ’Marine’, ’Natural Landscape’, ’Natural Patterns’, ’Animals’, ’Plants and Trees’, ’Water’, ’Seascape’, or ’Still-life’ then the artwork is considered to be Biophilic in nature else it is categorised as not significantly Biophilic. The gallery has been classified using the most dominant labels of the artwork.  Users can search artworks by visiting respective galleries or by simply searching with the individual labels. 

\section{Conclusion and Future Work}\label{sec:con}
Biophilic art can stimulate a sense of connection between the observer and nature or natural elements. The goal of this study has been to create and facilitate the recognition and accessibility of Biophilic art. A review of the literature revealed that this is important for homes and workplaces that have limited access to the natural environment. Hence, a well-curated collection of Biophilic artworks displayed through Textura Digital Canvas has been created that can be used to create a healthier and more Biophilic environment. Given previous studies supporting the impact of Biophilic art and design on well-being, it is hoped that the curation of a Biophilic collection will benefit positive emotions and mitigate stress. Compared to other methodologies, the techniques used in this work produced more accurate classification results, outperforming popular methods for multi-label classification. This may be mainly due to the inclusion of the CLIP image encoder \cite{3}, which was trained on 400 million images from more than 1000 classes, improving the extraction of characteristics from various types of images. This study is the first of its kind to use such techniques to classify and curate a collection of art based on Biophilic features. The resulting classifier model is a shift from the conventional methods of manually curating artworks and will benefit further work in Biophilic artwork and design. Moreover, using state-of-the-art image recognition and classification techniques, this work elucidates the unique patterns and themes that underlie Biophilic traits. This supports a novel approach to analysing and curating artworks based on their Biophilic aspect. Our well-curated Biophilic dataset can not only be used for curating public spaces but can also be used in research across several disciplines, such as the study of Biophilia on health and wellbeing. Future work will further investigate how the art classified as Biophilic using artificial intelligence could benefit psychological function. In addition, we will create algorithms to build a personalised system that could recommend Biophilic artworks based on the emotional response of a user, the circadian cycle, individual differences in environmental conditions in user states and traits, as well as environmental factors, etc. In the realm of artistic practice, our work will be useful in providing creatives with a non-human analysis of the Biophilic capacity of their work. Further studies will explore the feasibility of generating novel artwork using the underlying patterns learnt by our classifier.

\section{Declarations}
\subsection{ Availability of materials and data}
All data used for this research is available at: \\ \url{https://www.kaggle.com/datasets/purnakar/biophilic-artwork-dataset}. 

For reproducing our results, the source code is publicly available at: \\ \url{https://github.com/purnakar18/Biophilic-Artwork-Classification}

\bibliographystyle{abbrv}
\bibliography{references}  

\end{document}